\documentclass[runningheads]{llncs}
\usepackage{graphicx}
\usepackage[T1]{fontenc}

\usepackage[utf8]{inputenc}
\usepackage{lscape}
\usepackage{makecell}

\begin{document}
\title{Operational Collective Intelligence of Humans and Machines}
\titlerunning{Operational Collective Intelligence}
%
\author{
Nikolos Gurney\inst{1}\orcidID{0000-0003-3479-2037} \and
Fred Morstatter\inst{3} \and
David V. Pynadath\inst{1,2}\orcidID{0000-0003-2452-4733} \and
Adam Russell\inst{3} \and
Gleb Satyukov\inst{3}}
\authorrunning{N. Gurney et al.}
%
\institute{Institute for Creative Technologies, University of Southern California, 90094 USA 
\url{http://ict.usc.edu/}\\
\email{\{gurney,pynadath\}@ict.usc.edu}\and
Computer Science Department, University of Southern California, 90007 USA\and
Information Sciences Institute, University of Southern California, 90292 USA 
\url{http://www.isi.edu//}\\
\email{\{fredmors,arussell,gleb\}@isi.edu}}

\maketitle
\begin{abstract}
We explore the use of aggregative crowdsourced forecasting (ACF) \cite{benjamin2023hybrid,morstatter2019sage} as a mechanism to help operationalize ``collective intelligence'' of human-machine teams for coordinated actions. We adopt the definition for Collective Intelligence as: ``A property of groups that emerges from synergies among data-information-knowledge, software-hardware, and individuals (those with new insights as well as recognized authorities) that enables just-in-time knowledge for better decisions than these three elements acting alone.'' \cite{suran2020frameworks} Collective Intelligence emerges from new ways of connecting humans and AI to enable decision-advantage, in part by creating and leveraging additional sources of information that might otherwise not be included. Aggregative crowdsourced forecasting (ACF) is a recent key advancement towards Collective Intelligence wherein predictions (X\% probability that Y will happen) and rationales (why I believe it is this probability that X will happen) are elicited independently from a diverse crowd, aggregated, and then used to inform higher-level decision-making. 
This research asks whether ACF, as a key way to enable Operational Collective Intelligence, could be brought to bear on operational scenarios (i.e., sequences of events with defined agents, components, and interactions) and decision-making, and considers whether such a capability could provide novel operational capabilities to enable new forms of decision-advantage. 

\keywords{First keyword  \and Second keyword \and Another keyword.}
\end{abstract}
\section{Introduction}

Collective Intelligence (CI) emerges from new ways of connecting humans and machines to enable decision-advantage, at least in part, by creating and leveraging additional sources of information that decision-makers might otherwise not include \cite{atanasov2017distilling,budescu2006confidence,mellers2014psychological}. Operational scenarios present a unique challenge for typical collective intelligence approaches. The wisdom of the crowd, the archetypal collective intelligence, works in classic applications, like predicting the weight of a cow, because the probability distributions of responses in these instances have median estimates centered near the ground truth, e.g., the weight of the cow \cite{galton1907vox,surowiecki2005wisdom}. Typical operational scenarios necessitate a different approach because the ``crowd'' is often heavily biased. For example, in a military setting operators at various levels are privileged to unique knowledge, some of which may alter their predictions in an adversarial reasoning scenario. Moreover, operational scenarios increasingly rely on input from machine intelligence, which decision-makers must integrate with human judgments. Aggregative crowdsourced forecasting (ACF) is a recent, pivotal advancement in CI methods. ACF helps decision-makers from commercial companies, government organizations, and militaries overcome the intractable problem of individual bias by collecting many predictions and integrating the best of machine intelligence into the decision process. To date, ACF has been used for risk management at a strategic level, focusing on tasks like informing policy, anticipating instability, balancing research portfolios, identifying emerging technology, and serving as an early warning mechanism for decision-making. 

This document provides a preliminary exploration and evaluation of the potential use of aggregative crowdsourced forecasting as a mechanism to help operationalize ``collective intelligence'' for operational scenarios. After operationalizing pertinent terminology and laying out the critical research questions, we identify gaps in the research and propose ways of empirically testing solutions to the open questions. 

\section{Background}
The insight that a group of forecasters may be more reliable than a single individual, colloquially known as \textit{the wisdom of a crowd}, is far from new. It is in Aristotle's \textit{Politics} \cite{landemore2012collective} and was famously championed by the British polymath Sir Francis Galton. (Both use the accuracy achieved by statistical aggregation of forecasters as motivation for democracy; Aristotle waxed poetic about art judgments while Galton, more practically, used the statistical accuracy of a crowd's guesses of the dressed weight of a cow to argue that popular judgments are valuable to societies \cite{galton1907vox}.) Modern technology has undeniably changed the dynamics of such crowd wisdom. Although digital technology facilitates aggregation of and eases access to crowd wisdom \cite{kameda2022information}, it can also undermine it because crowd wisdom hinges on natural variation in the forecasters' information. Individual forecasters who use digital technology to change their knowledge of other forecasters' knowledge may become biased and reflect their new knowledge in their forecasts, ultimately undermining the crowd's wisdom \cite{lorenz2011social}. Studying the process of crowd decision-making and how the crowd's wisdom emerges may facilitate mitigation, if not reversal, of digital technology's undermining effects on the crowd wisdom \cite{levy1997collective,suran2020frameworks}. 

\subsection{Collective Intelligence}
Collective intelligence (CI) is a ``form of universally distributed intelligence, constantly enhanced, coordinated in real-time, and resulting in the effective mobilization of skills'' \cite{levy1997collective}. We are primarily interested in collective intelligence that emerges from new ways of connecting humans and AI to enable decision advantage in instances of adversarial reasoning. Decision advantages are made possible by connections that allow the collective to create and leverage information sources that decision-makers might otherwise not include in a decision process. Adversarial reasoning involves ``determining the states, intents, and actions of one's adversary, in an environment where one strives to effectively counter the adversary's actions.'' \cite{kott2015toward} 

Suran, Pattanaik, and Draheim introduce a convenient, generic framework for conceptualizing CI \cite{suran2020frameworks}. 
They posit that a CI is composed of staff with a goal they are motivated to achieve through some definable process. The ``crowd'' in this instance is the active staff members who contribute to the CI (some staff in a CI may be passive, e.g., beneficiaries). The CI goal is simply a desired outcome---for example, anticipating an opponent's strategy or recognizing deceptive tactics. Although it is common to think of the goal as being community-driven, individuals may also participate in a CI to achieve their own unique goals: A software engineer may commit updates to an open-source repository to improve its utility for their own project but in so doing advance the collective goal of having a robust, stable resource for a community of users. Since the authors excluded AI systems, motivation is simply the \textit{why} behind contributions to the collective, be it intrinsic (e.g., a passion for the cause) or extrinsic (e.g., money). Lastly, the definable process of a CI describes the staff's interactions in pursuit of the goal---think of the process as the \textit{how} that describes the way in which a CI achieves intelligence. 

Each aspect of the Suran, Pattanaik, and Draheim framework builds upon extensive research---they document more than 9,000 publications in the space \cite{suran2020frameworks}. We are interested in the process---the how---of CI. Unsurprisingly, many publications in this space develop, review, and refine different CI processes. The project they lean on to develop the process aspect of their generalized framework looked at argument mapping tools, which diagram evidence for and against a cause, to improve online collective efforts \cite{iandoli2009enabling}. Proper deployment of these decision support systems can improve the process of CI (i.e., lead to better outcomes) by helping staff understand the interactions that transpire during collective decision-making. Of course, decision support systems are continually improving---machine learning and artificial intelligence technologies are dramatically enhancing their capabilities. Contemporary technologies facilitate mapping interactions and enhancing the information generated by those interactions \cite{morstatter2019sage}. Such technologies are paving the way for operational collective intelligence.

\subsection{Aggregative Crowdsourced Forecasting}
We define aggregative crowdsourced forecasting (ACF) as a type of forecasting wherein predictions (X\% probability that Y will happen) and rationales (why I believe it is this probability that X will happen) are elicited independently from a diverse crowd and aggregated into a single estimate for informing higher-level decision-making. Aggregation is carried out via machine learning models capable of identifying unique data features that may lead to biased forecasts, accounting for these features, and adjusting the weight of individual forecasts to produce a more reliable crowdsourced estimate. 

Conceptualized in this way, ACF is a type of intelligence that plays on the relative strengths of each forecaster to achieve otherwise unobtainable outcomes \cite{benjamin2023hybrid,dellermann2019hybrid,rafner2021revisiting}. Key features of this type of intelligence include humans and machines working together towards a collective goal, both agent types continually learning and improving, and the ability to achieve superior solutions than either type alone could realize \cite{dellermann2019hybrid}. A core strength of ACF is the machine models and human experts bringing unique skills to bear on their collective goal. The statistical models underpinning machine intelligence are able to accurately and efficiently identify patterns in structured data, which empowers them to make predictions that may avoid human biases. On the other hand, human experts are not reliant on structured data, which empowers them to see features in data that may go unnoticed by machine models. Together, the unique skills of machines and humans allow such an ACF to achieve superior performance.

\section{Operational Collective Intelligence}
Military, business, and other structured organizations rely on definable common occurrences, known as operational scenarios, to manage people, resources, and processes. Operational scenarios describe sequences of events using a set of defined agents, components, and interactions. Importantly, the agents, whether human or intelligent machines, that eventually carry out organizational processes based on operational scenarios are often privy to unique information that may bias their decision making and behavior. This feature can undermine the strengths of typical approaches to collective intelligence. Consider a military scenario in which field operators are intimately aware of situational factors that command personnel cannot observe while command personnel are privy to classified mission details that they cannot share with operators. Applying ACF, which is able to play on the strengths and weaknesses of different agents, to such operational scenarios may enable \textit{Operational Collective Intelligence} (OCI). 

\subsection{Research Questions}
Addressing several key, basic research questions will help us assess the viability of OCI, the main questions being whether and to what degree ACF has the potential for substantive operational impact. One of the most refined examples of an ACF is the Synergistic Anticipation of Geopolitical Events (SAGE) system \cite{benjamin2023hybrid}, which was developed as part of the IARPA Hybrid Forecasting Competition \cite{IARPA_2015}. SAGE is a hybrid forecasting system that gives forecasters the ability to combine their own judgments with model-based solutions. Its development emphasized making verifiable probabilistic predictions while being domain agnostic. Importantly, empirical results show that SAGE can result in improved aggregate forecasts. One insight is that forecasters needed a base level of expertise to make the most of the inputs provided by SAGE. This observation reinforces previous findings related to crowd-based forecasts \cite{atanasov2017distilling}. A unique feature of SAGE is its ability to facilitate qualitative (e.g., will a given candidate win a national election) and quantitative questions (e.g., what percent of the vote will each candidate receive). 

Machine-aided Analytic Triage with Intelligent Crowd Sourcing (MATRICS) is also a system developed under the IARPA Hybrid Forecasting Competition \cite{huber2019matrics}. Like SAGE, MATRICS combines human and machine predictions to generate a hybrid forecast. It similarly provides human forecasters guidance from a machine intelligence intended to improve their judgments before a final aggregation stage that mixes human and machine forecasts. MATRICS and SAGE differ from other machine-human hybrids in that they explicitly work to integrate wisdom-of-the-crowd models in their computation. Although neither system was explicitly developed as a solution for operational scenarios, the success of both suggests that they could be developed into relevant operational tools. 

An earlier system, CrowdSynth, also sought to leverage the relative strengths of human and machine intelligence by having an AI decide how to combine multiple contributions \cite{kamar2012combining}. CrowdSynth was able to optimize engagement with citizen scientists who contributed to a constellation classification task. It accomplished this by balancing expected votes from contributors with a model of what it believed would be the ultimate consensus classification. The interactive intelligence of MATRICS and SAGE differentiates them from CrowdSynth---they not only consider the wisdom of the crowd, but work directly with it to improve the crowd's performance. Although CrowdSynth was able to reduce the number of contributors needed for an accurate classification, it did not have this collective feature. 

If we step away from the aggregative, interactive features of SAGE and MATRICS, there are numerous \textit{hybrid intelligence} systems \cite{dellermann2019hybrid}. For example, Dong and coauthors developed a hybrid intelligence algorithm for E-commerce sales volume forecasting \cite{dong2022human}. Their simulations suggest that access to the AI system is not always helpful in a purely collaborative scenario---particularly when the humans do not trust the system. Interestingly, more accurate forecasting occurred when the AI was an opponent in a competitive scenario. Another example from Wu et al. uses models of human cognition and examples from human designers, rather than actual humans, to aid an automotive design AI system \cite{wu2023human}. They demonstrate that the system is superior to a baseline model that lacks the human model plus knowledge integration and argue that it is capable of exceptional creativity. In yet another example, Russakovsky, Li, and Fei-Fei developed a model that combined multiple algorithmic and human annotations of images to improve overall classification \cite{russakovsky2015best}. Although these systems have apparent utility in operational scenarios, including those that require adversarial reasoning, they are limited in that they do not aggregate and/or account for the knowledge, experience, and seniority of human staff. This core feature of ACF is what we believe will set it apart as an operational collective intelligence. However, even though experimentally validated models exist in other domains, there are still pertinent questions to raise about the viability of ACF for operational scenarios.

\subsubsection{What are the determinants of ACF's practicality for operational scenarios?}
Aggregative Crowdsourced Forecasting is a data-centric process. The limited amount of ACF research makes formal estimation of data requirements untenable, however, related research points to some guidelines. Training the core machine learning models not only requires more data than is typical for other statistical models, but the data also need to be easy to access, well-structured, and relatively errata free \cite{bollier2010promise,jordan2015machine}. Each element needed to construct a functional ACF will have these requirements. It follows that access to high-quality, relevant training data is a key determinant of successfully deploying ACF in an operational scenario. The very nature of adversarial scenarios means that troves of valuable data are collected and maintained. For example, in a military command and control operational scenario, an effective ACF model will rely on data including features such as the opponent's plans for and perceptions of friendly forces; the opponent's doctrine, tactics, and training; environmental constraints from terrain to weather; capital resources such as ammunition, artillery, communications, personnel, etc.

Access to high-quality data is only half of the modeling task---the modeling technology needs to match the challenge as well. Modeling technology for crowd-sourced forecasting has only recently become broadly viable \cite{hassani2015forecasting}. In geopolitical modeling, for example, robust forecasting models emerged roughly when access to big data and high-powered computing were democratized \cite{leigh2006competing}. Additionally, even though combining forecasts from different sources, such as human and machine intelligence, is not new, the necessary computational power and algorithm technologies for large-scale modeling are fairly recent developments \cite{wang2023forecast}. The cost of developing, maintaining, and deploying such models means that early ACF models will most likely be applied in settings where the risk of failure and cost of failure are uniquely high. 

The need for extant, reliable data and established, state-of-the-art forecast modeling currently limits the operational scenarios for which ACF is a viable instrument. In the cases of SAGE \cite{benjamin2023hybrid} and MATRICS \cite{huber2019matrics}, the data were determined and provided by the sponsor of the competition (IARPA \cite{IARPA_2015} plus they were adapted for a well-studied forecasting domain--geopolitical events. Pursuing additional work in this area would undoubtedly allow for model improvement. However, working with new data in other domains may foster faster, more reliable development \cite{jordan2015machine}. One domain with well-defined, studied, and documented operational scenarios is military decision-making \cite{svenmarck2018possibilities}. And even though much of military operational data is not publicly available, many military organizations allow researchers to publish appropriately cleaned data to gain access to state-of-the-art machine learning \cite{galan2022military}, meaning it may prove a viable domain for developing OCI. Other potential domains exist, such as business intelligence, but access issues commonly fetter these data too \cite{peled2013politics,pencheva2020big}. 

\subsubsection{What are the key human factors of a successful OCI?}

The goal of operational collective intelligence is to provide ultimate decision-makers with guiding input based on the best of human and machine intelligence. An important aspect of the \textit{how} of OCI is the human staff's response to the machine intelligence. Historically, people preferred human over algorithmic judgment (i.e., machine or artificial intelligence) \cite{dawes1989clinical,dietvorst2015algorithm}. Whenever and wherever this tendency still exists, it may challenge OCI. Fortunately, algorithm aversion is not a foregone conclusion of human behavior---increasingly, researchers are observing instances of the opposite, that is, people seeking out algorithmic input \cite{logg2019algorithm}. Unfortunately, people occasionally put too much trust in algorithmic decision aids \cite{parasuraman1997humans,wang2016trust}, which may also challenge OCI. Thus, a pivotal human factor of a successful OCI is well-calibrated trust in its machine intelligence. A well-calibrated decision-maker will know when to heed and ignore an OCI's input. From a design perspective, this reality suggests a need for implementing tools to measure and predict trust in the OCI \cite{gurney2022measuring,gurney2023comparing} along with ways to calibrate that trust. The trust calibration literature is quite broad, but some obvious, simple steps include engineering the messages such that they guide users to the appropriate level of trust \cite{wang2016trust} and providing interpretable explanations of the system's output \cite{pynadath2022explainable,zhang2020effect}. 

Extensive evidence across a variety of domains supports the observation that a crowd can outperform experts \cite{galton1907vox,landemore2012collective,surowiecki2005wisdom}. Nevertheless, expertise still matters: Clever aggregation that weighs individual contributors' forecast relative to their knowledge can yield even wiser crowds \cite{budescu2006confidence,budescu2015identifying,mannes2014wisdom,welinder2010multidimensional}. For example, in an image annotation setting where multiple annotators labeled images, modeling individual annotators as multidimensional entities with competence, expertise, and bias variables resulted in better aggregation of their opinions \cite{welinder2010multidimensional}. A similar result was observed in studies that asked participants to forecast current events and economic outcomes \cite{budescu2015identifying}. 

Situational factors, however, may undermine the effect of individual expertise. One study asked participants to estimate corporate earnings and found that having access to public information resulted in participants under-weighting their own private information \cite{da2020harnessing}. This behavior resulted in individual estimates that were more accurate but crowd estimates that were less accurate---a result that underscores how important aggregation that accounts for human factors is to an OCI.

\subsubsection{What are the key machine intelligence factors of a successful OCI?}
The machine intelligence (i.e., algorithmic) factors of a successful OCI are subject to the same limitations as machine learning more generally: they perform best on well-defined tasks \cite{christiano2017deep}, when the environment is simple and stable \cite{brynjolfsson2017can}, and when high-quality training data are available \cite{ratner2017statistical}. Operational scenarios are driven by the need for predictable, defined protocols for common occurrences, meaning that the task is generally well-defined for the machine intelligence that underpins ACF systems. However, one factor that could undermine them could also make OCI invaluable: the ability to handle exceptional task cases. Thus, a critical stage in developing OCI is engineering pathways into the ACF system for handling exceptions. One solution is to empower the ACF model with a classification tool for detecting exceptions \cite{haixiang2017learning} and passing exceptional cases off to human staff, who are typically better at handling them \cite{benjamin2023hybrid}. 

Using operational scenarios is most common in stable, i.e., predictable, environments. However, stable and predictable does not necessarily mean an operational scenario is simple. Many operational scenarios exist as responses to complicated environments in which simplification can improve a given operation. Machine intelligence, at least historically, is not considered an applicable solution for tasks in complex settings, although given sufficient data and training resources, models are able to generalize their learning (e.g., \cite{mnih2015human}). Innovative modeling approaches in the space of neural-symbolic reasoning, however, may prove a solution for ACF in complex environments \cite{garcez2022neural}. 

Access to high-quality, operation-relevant data is critical for training an ACF model. Data quality is simply the degree to which archived training data reflects the real world \cite{heinrich2018requirements}. Moreover, a training data set's relevance is contingent on a given organization and the operational setting. For example, network intrusion detection is complicated by the fact that operational data from one network setting is rarely relevant to another, which creates a lack of training data \cite{sommer2010outside}. Although network data from other operational settings could be used to train a model, it likely will not reflect ``the real world'' in new network security settings, meaning it is poor-quality data. The impact of poor data quality on a given model is contingent on the type of model and the task at hand \cite{budach2022effects}---if the cost of errors is low, then poor quality data may be a reasonable tool for getting a model established to facilitate collection of more relevant data \cite{sommer2010outside}. In the case of ACF, data quality may be the result of how the crowd was selected \cite{li2014wisdom,li2015cheaper}. This suggests that an OCI may benefit from tools that allow it to select which training data are used for a given operational scenario. 

ACF models are subject to the same generalization challenges as other machine learning models. The models underpinning ACF typically adopt the classic statistical assumption of independent and identically distributed data (i.i.d.)---a very strong assumption that is made necessary by the intractable complexity of its absence. Out-of-Distribution (OOD) data, i.e., real-world data, can prove a significant challenge for these models. ML models for OOD data are still in their nascent phase, but improving \cite{liu2021towards}, including in forecasting applications \cite{shoeibi2024automated}. Models that are able to handle OOD data that do not abide by the i.i.d. assumption will also prove a key MI factor of successful OCI. 

\subsection{Potential Operational Collective Intelligence Impact(s)}
Research already documents many impacts of human-machine collective intelligence. For example, one application in a healthcare setting demonstrated that significant error reduction is possible by comparing single medical diagnosticians and a CI system---the diagnosticians achieved 46\% accuracy versus the system's 76\% \cite{kurvers2023automating}. A famous, but perhaps underappreciated, example of collective intelligence is Google. It relies on judgments and decisions from people worldwide to refine its representations of content on the web and improve algorithmic recommendations \cite{malone2010collective}. The general impact of Google on individual decision-makers, collectives, and society at large is undeniable (and likely impossible to estimate). Similarly, enumerating the unique ways in which it has changed the course of human history is probably impossible. 

Military, business, and other organizations define operational scenarios to improve operational outcomes. Outcome improvements result from using operational scenarios during training and simulation settings, which aids staff in learning what decision features to focus on, and using them during actual operations, which may aid staff in identifying anomalous features of an operation. We envision OCI improving organizations' decision accuracy by improving their ability to listen to the most knowledgeable staff, adapt to unpredictable scenarios by recognizing situations in which machine intelligence is underspecified, and increase decision speed by leveraging the system's ability to process new data relative to human staff alone. In an adversarial reasoning scenario, whether in military, business, or political settings, such an increase in decision time could allow an organization to gain a competitive advantage.

\section{Discussion}
Research documents impressive outcomes of combining human and machine intelligence, i.e., Collective Intelligence, to solve complex operational problems\cite{suran2020frameworks}. Significant advances in machine learning have facilitated aggregative crowdsourced forecasting, a CI capable of taking the best input from individual staff members of a crowd and intelligently guiding decisions \cite{benjamin2023hybrid,morstatter2019sage}. Applying this technology to operational scenarios, we believe, is a way to realize Operational Collective Intelligence.  

\subsection{Advantages}
OCI's main advantage lies in its ability to integrate human and machine intelligence while applying lessons learned from crowd wisdom quickly and accurately. Human decision-makers are simply not capable of considering troves of data or agnostically weighing input from a large number of forecasters in the same fashion. For example, a robust OCI could consider seasonal variations in the weather, such variations' impacts on operators' well-being, and the cost of delays without referring to outside resources in minutes if not seconds. It is not unreasonable for a human decision-maker to rely on a cadre of workers for each sub-prediction, which might require hours or days, and then when they integrate the various outputs, allow their own biased judgment to impact how they weigh the input from different forecasting teams.  

\subsection{Practicality}
The key to OCI's practicality is data. The lock is well-designed machine intelligence. Without these, OCI may prove more of a hindrance than a benefit. ACF models are only as good as the data used to train them. Current state-of-the-art ACF models require more data than is typical for other statistical models, underscoring the importance of having ready access to well-structured, errata-free data \cite{bollier2010promise,jordan2015machine}. Without such data, OCI is not practical. Additionally, although considerable effort has already gone into designing and validating ACF models, such as what is reported in \cite{benjamin2023hybrid}, much work is still needed to ensure the practicality of an OCI that hinges on ACF. 

The staff interacting with an OCI and their responses to it are of similar importance. Despite historical evidence of people being averse to algorithmic input\cite{dietvorst2015algorithm}, there is increasing evidence that people not only support but seek out algorithmic aid \cite{logg2019algorithm}. With this increased willingness to rely on algorithmic decision aids like OCI comes an increased need to ensure that the human staff of an operation is well-calibrated to its use \cite{pynadath2022explainable,wang2016trust}.

\section{Conclusion}
Aggregative crowdsourced forecasting helps decision-makers overcome intractable forecasting problems, such as individual biases, by collecting many predictions and integrating the best of machine intelligence into the decision process. Operational scenarios are an obvious application of this recent advancement in collective intelligence, but translating existing ACF models for operational scenario application is not straightforward. Data, modeling, and human factors make this translation a challenge. ACF is a data-centric process in which each element requires readily available data that are well-structured and relatively errata-free. Fortunately, there are many operational applications, such as adversarial scenarios, in which troves of such data are collected. ACF also hinges on state-of-the-art forecasting models that require considerable computational resources. Beyond the data and computational requirements are the human factors of operational scenarios, from how the human staff responds to machine intelligence to their relative expertise. Addressing these challenges will enable ACF to unlock operational collective intelligence, which we believe will prove a pivotal advancement in the coordinated action of human-machine teams.

\begin{credits}
\subsubsection{\ackname} The project or effort depicted was or is sponsored by the U.S. Government under contract number W911NF-14-D-0005. The content of the information does not necessarily reflect the position or the policy of the Government, and no official endorsement should be inferred.

\subsubsection{\discintname}
The authors have no competing interests to declare that are relevant to the content of this article.
\end{credits}

\bibliographystyle{splncs04}
\bibliography{bib}

\end{document}